\documentclass[letterpaper]{article} 
\usepackage{aaai25}  
\usepackage{times}  
\usepackage{helvet}  
\usepackage{courier}  
\usepackage[hyphens]{url}  
\usepackage{graphicx} 
\usepackage{natbib}  
\usepackage{caption} 
\usepackage{algorithm}
\usepackage{algorithmic}

\usepackage{algorithmic}
\usepackage{bm}
\usepackage{amsmath}
\usepackage{amssymb}
\usepackage{bm}
\usepackage{multirow}
\usepackage{xcolor}         

\usepackage{cleveref}
\usepackage{booktabs} 
\usepackage{newfloat}
\usepackage{listings}
\usepackage{multirow}
\usepackage{xcolor}         
\usepackage{cleveref}

\usepackage{newfloat}
\usepackage{listings}
\DeclareCaptionStyle{ruled}{labelfont=normalfont,labelsep=colon,strut=off} 
\floatstyle{ruled}
\newfloat{listing}{tb}{lst}{}
\floatname{listing}{Listing}
\title{BloomScene: Lightweight Structured 3D Gaussian Splatting for \\ Crossmodal Scene Generation}

\author{
    Xiaolu Hou\textsuperscript{\rm 1}\equalcontrib, 
    Mingcheng Li\textsuperscript{\rm 1}\equalcontrib, 
    Dingkang Yang\textsuperscript{\rm 1, \rm 6}\thanks{Corresponding authors.}, 
    Jiawei Chen\textsuperscript{\rm 1}, 
    Ziyun Qian\textsuperscript{\rm 1}, 
    Xiao Zhao\textsuperscript{\rm 1}, \\
    Yue Jiang\textsuperscript{\rm 1}, 
    Jinjie Wei\textsuperscript{\rm 1}, 
    Qingyao Xu\textsuperscript{\rm 1}, 
    Lihua Zhang\textsuperscript{\rm 1,\rm 2,\rm 3,\rm 4,\rm 5\textdagger}
}
\affiliations{
    \textsuperscript{\rm 1} Academy for Engineering and Technology, Fudan University\\
    
    \textsuperscript{\rm 2} Institute of Metaverse \& Intelligent Medicine, Fudan University\\
    
    \textsuperscript{\rm 3} Engineering Research Center of AI and Robotics, Ministry of Education\\
    
    \textsuperscript{\rm 4} Jilin Provincial Key Laboratory of Intelligence Science and Engineering \\

     \textsuperscript{\rm 5}  Artificial Intelligence and Unmanned Systems Engineering Research Center of Jilin Province\\

     \textsuperscript{\rm 6}  ByteDance Inc\\

     xlhou23@m.fudan.edu.cn, mingchengli21@m.fudan.edu.cn

}

\usepackage{bibentry}

\begin{document}

\maketitle

%

\begin{abstract}
With the widespread use of virtual reality applications, 3D scene generation has become a new challenging research frontier. 3D scenes have highly complex structures and need to ensure that the output is dense, coherent, and contains all necessary structures. Many current 3D scene generation methods rely on pre-trained text-to-image diffusion models and monocular depth estimators. However, the generated scenes occupy large amounts of storage space and often lack effective regularisation methods, leading to geometric distortions.
To this end, we propose BloomScene, a lightweight structured 3D Gaussian splatting for crossmodal scene generation, which creates diverse and high-quality 3D scenes from text or image inputs. 
Specifically, a crossmodal progressive scene generation framework is proposed to generate coherent scenes utilizing incremental point cloud reconstruction and 3D Gaussian splatting.
Additionally, we propose a hierarchical depth prior-based regularization mechanism that utilizes multi-level constraints on depth accuracy and smoothness to enhance the realism and continuity of the generated scenes.
Ultimately, we propose a structured context-guided compression mechanism that exploits structured hash grids to model the context of unorganized anchor attributes, which significantly eliminates structural redundancy and reduces storage overhead.
Comprehensive experiments across multiple scenes demonstrate the significant potential and advantages of our framework compared with several baselines.

\end{abstract}

%
\begin{links}
    \link{Code}{https://github.com/SparklingH/BloomScene}
\end{links}

\section{Introduction}
Currently, there is a growing demand for 3D content in virtual reality. However, creating 3D content is time-consuming and requires deep expertise, making 3D content generation a challenging frontier. 
In the 2D domain, sufficient annotated datasets have greatly contributed to the development of text-to-image generation models \cite{rombach2022high}, enabling users to generate images through natural language. However, the shortage of annotated 3D datasets limits the application of supervised learning in 3D content generation \cite{ouyang2023text2immersion}. 
To address this challenge, recent studies \cite{poole2022dreamfusion, lin2023magic3d} extract 2D priors from diffusion models through a time-consuming distillation process to optimize the generation of 3D content. However, these methods \cite{wang2024prolificdreamer} have limitations when extended to fine-grained scenes with outward-facing viewpoints. 
Therefore, several methods \cite{hollein2023text2room, ouyang2023text2immersion, hou2024sceneweaver} that combine pre-trained text-to-image generation models \cite{rombach2022high} with monocular depth estimators \cite{bhat2023zoedepth, ranftl2020towards} are receiving increasing attention due to their advantages in complex 3D scene generation.

Some methods \cite{hollein2023text2room, fridman2024scenescape} generate 3D indoor scenes represented by mesh using a progressive framework but are prone to distorted or over-smoothing regions when applied to outdoor scenes. With the wide application of NeRF \cite{mildenhall2021nerf} in novel view synthesis tasks, Text2NeRF \cite{zhang2024text2nerf} generates 3D scenes represented by NeRF with a progressive framework. Although this method can generate high-quality scenes, the generation time is still quite long.
Recently, 3D Gaussian Splatting (3DGS) \cite{kerbl20233d} has been widely used for high-quality scene generation due to its excellent generation quality and real-time rendering capabilities. Among them, LucidDreamer \cite{chung2023luciddreamer} and Text2Immersion \cite{ouyang2023text2immersion} use a progressive generation framework that follows the optimization goals of 3DGS to achieve domain-free 3D scene generation. 
Although previous 3DGS-based approaches have made some progress in 3D scene generation, they still suffer from the following limitations: ({\romannumeral1}) Rely only on photometric loss in the scene optimization process, lack sufficient regularization techniques, and are prone to artifacts and ambiguities.({\romannumeral2}) 3DGS requires millions of 3D Gaussians to represent each scene, resulting in high memory requirements, increasing storage costs, and end-device burden.

To address the above problems, we propose BloomScene, a lightweight structured 3D Gaussian splatting for high-quality crossmodal 3D scene generation. BloomScene has the following three core contributions. 
(\romannumeral1) We propose a crossmodal progressive scene generation framework for generating 3D scenes via progressive point cloud reconstruction and 3D Gaussian splatting. 
(\romannumeral2) Additionally, a hierarchical depth prior-based regularization mechanism is proposed to enhance the realism and continuity of the scene by implementing multi-level depth accuracy constraints and smoothness constraints.
(\romannumeral3) We propose a structured context-guided compression mechanism, which leverages a structured hash grid to model the context of unorganized anchor attributes, thus sufficiently compressing the model storage space.
Comprehensive experiments demonstrate that the scenes generated by our framework significantly outperform baselines in terms of fidelity and geometric consistency, proving its significant potential and advantages in complex 3D scene generation.

\begin{figure*}[t]
\centering
\includegraphics[width=0.8\textwidth]{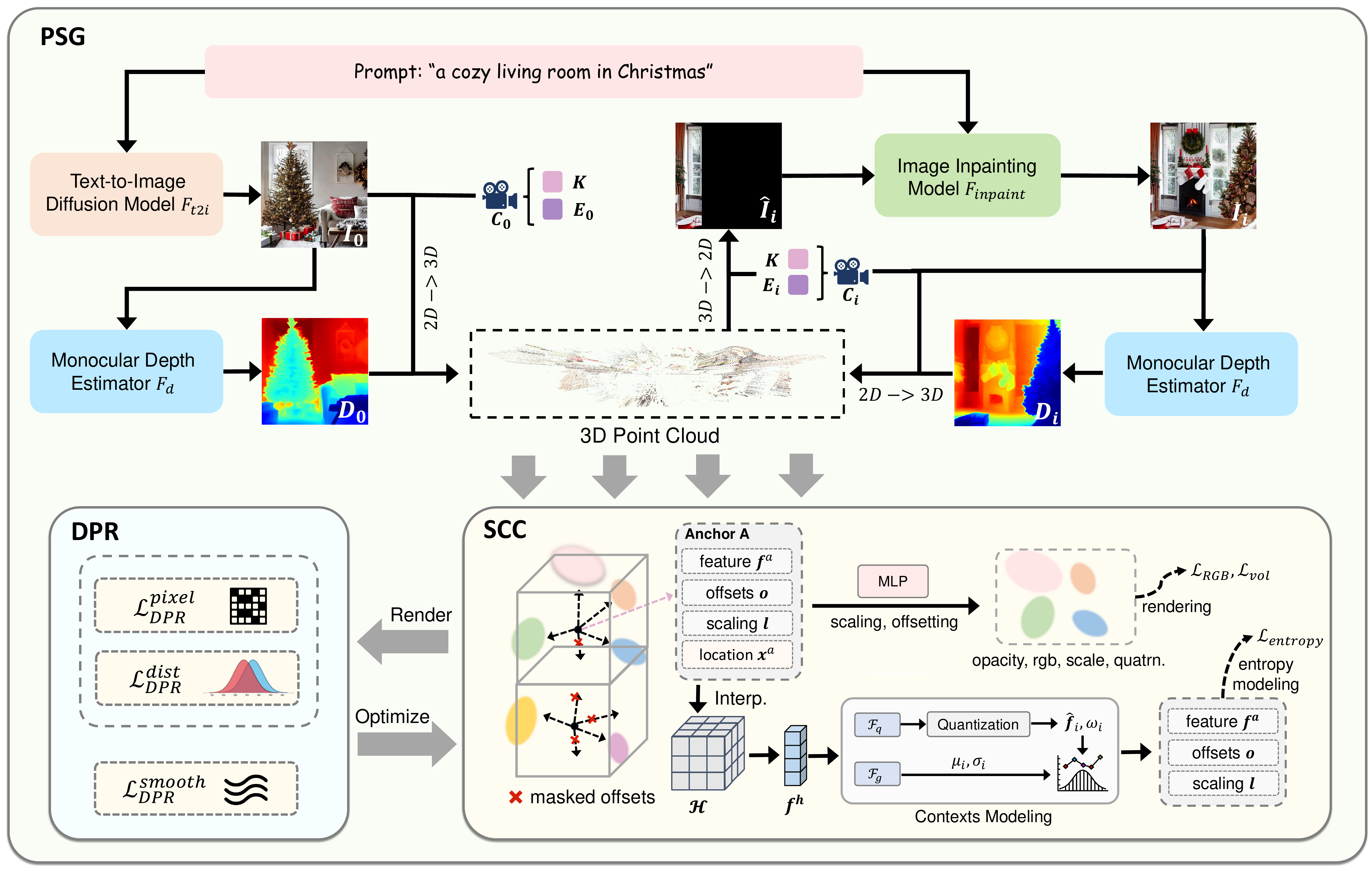}
\caption{The overall framework of the proposed BloomScene. 
BloomScene utilizes the proposed crossmodal Progressive Scene Generation (PSG) framework to generate 3D scenes from the text prompts progressively. Moreover, the hierarchical Depth Prior-based Regularization (DPR) mechanism is applied to the 3DGS to enhance the realism and continuity of the generated scenes. Eventually, Structured Context-guided Compression (SCC) is employed to mine structural correlations in 3DGS and reduce storage overhead.}
\label{fig:framework}
\end{figure*}

\section{Related Work}

\noindent\textbf{Crossmodal 3D Scene Generation.}
Generating 3D content through language enables users to realize their demands without modeling skills. Existing methods \cite{mohammad2022clip, lee2022understanding, poole2022dreamfusion, lin2023magic3d, wang2024prolificdreamer, tang2023dreamgaussian} optimize 3D content using the prior knowledge of pre-trained models \cite{radford2021learning, rombach2022high}. While progress has been made in the single-object generation, ensuring texture and structure consistency is still difficult when generating complex scenes with outward-facing viewpoints \cite{wang2024prolificdreamer}. The application of diffusion models in image inpainting has driven the development of progressive scene generation \cite{fridman2024scenescape, hollein2023text2room, chung2023luciddreamer, ouyang2023text2immersion, engstler2024invisible, yu2024wonderjourney} by combining a monocular depth estimator \cite{bhat2023zoedepth, ranftl2020towards} to update the scene. Among them, although LucidDreamer \cite{chung2023luciddreamer} and Text2Immersion \cite{ouyang2023text2immersion} achieve higher-quality scene generation results using 3DGS \cite{kerbl20233d}, they are prone to artifacts and ambiguities due to their reliance on photometric loss alone. Therefore, we propose the hierarchical depth prior-based regularization mechanism for multi-level regularization of 3DGS.

\noindent\textbf{Efficient 3D Scene Representation.}
In 3D content generation, choosing an appropriate 3D representation is crucial. The classical explicit representations \cite{munkberg2022extracting, berger2014state} provide intuitive geometric control and are suitable for simple scenes, but may face memory and rendering efficiency issues in complex scenes. 
Neural network-based implicit representations \cite{mildenhall2021nerf, muller2022instant, barron2022mip} improve expressiveness but still require a trade-off between speed and quality.  
3D Gaussian Splatting (3DGS) \cite{kerbl20233d} achieves fast rendering and high-quality output results, but high storage requirements impose an additional burden. For this reason, some methods focus on value \cite{fan2023lightgaussian, navaneet2023compact3d} or structure representation \cite{lu2024scaffold} to reduce the computational burden. However, structural redundancy or anchor independence issues lead to lower compression efficiency. 
To this end, we propose the structured context-guided compression mechanism, which utilizes a structured hash feature grid to achieve contextual modeling of anchor point attributes for further compression of 3DGS.

\section{Methodology}
\subsection{Preliminaries}
3D Gaussian Splatting (3DGS) \cite{kerbl20233d} introduces the 3D Gaussians as differential volumetric representations of radiance fields, allowing high-quality real-time novel view synthesis. 
A set of splats is initialized from the calibrated camera poses and the sparse point clouds produced through Structure-from-Motion (SfM) \cite{snavely2006photo}.
Each Gaussian is represented by position $\bm{\mu}$ and covariance matrix $\bm{\Sigma}$, denoted as  $G(\bm{x})=e^{-\frac{1}{2}(\bm{x}-\bm{\mu})^T \bm{\Sigma}^{-1}(\bm{x}-\bm{\mu})}$. 
The covariance can be decomposed from a scaling matrix $\bm{S}$ and rotation matrix $\bm{R}$, expressed as $\bm{\Sigma} = \bm{R}\bm{S}\bm{S}^T\bm{R}^T $ with $\bm{S}$.
To render the color, 3DGS further optimizes opacity and Spherical Harmonic (SH) coefficients, following the point-based differential rendering by rasterizing anisotropic splats with $\alpha$-blending, denoted as:
\begin{equation}
\label{z-depth}
    \hat{\bm{C}}=\sum_i^N \mathbf{\bm{c}}_i \bm{\alpha}_i \prod_j^{i-1}\left(1-\bm{\alpha}_j\right), \quad \hat{\bm{D}}=\sum_i^N \bm{d}_i \bm{\alpha}_i \prod_j^{i-1}\left(1-\bm{\alpha}_j\right),
\end{equation}
where $\bm{c}_i$ and $\bm{\alpha}_i$ denote the color and opacity of the Gaussian, and $\bm{d}_i$ is the z-axis of the points by projecting the center of 3D Gaussians $\bm{\mu}$ to the camera coordinate.

\subsection{Crossmodal Progressive Scene Generation}
Previous methods \cite{wang2024prolificdreamer} have made progress in the single-object generation, but it is difficult to ensure texture and structural coherence when generating complex scenes with outward-facing viewpoints. 
To realize crossmodal 3D scene generation, we propose a crossmodal Progressive Scene Generation (PSG) framework to incrementally generate lightweight and high-quality scenes with reference to previous work \cite{ouyang2023text2immersion, chung2023luciddreamer}. 
The main workflow of the proposed PSG is shown in \Cref{fig:framework}, which consists of four main phases: point cloud construction, 3D Gaussians initialization, hierarchical Depth Prior-based Regularization (DPR) mechanism to optimize the quality of the 3DGS-generated scenes, and Structured Context-guided Compression (SCC) for reducing the storage overhead. All these phases constitute the PSG framework for realizing cross-modal, high-quality 3D scenes.

\noindent\textbf{Point Cloud Construction.}
Given a text prompt $y$, our goal is to generate a 3D scene that matches $y$ in a crossmodal manner. We use a text-conditioned image inpainting model $F_{inpaint}$ and a monocular depth estimator $F_d$ to progressively inpaint and update the scene.
The pre-trained text-to-image diffusion model $F_{t2i}$ is used to generate the initial image $\bm{I}_0$ from text prompt $y$. If the input is an image without a corresponding text description, the pre-trained image-to-text generation model $F_{i2t}$ is used to generate the corresponding text prompt $y$ from $\bm{I}_0$, constituting an image-text pair. $F_d$ is then used to obtain the depth map $\bm{D}_0$ from $\bm{I}_0$.

Unlike previous work \cite{chung2023luciddreamer}, we choose to set the camera to rotate sequentially from the initial position to both sides to minimize the cumulative error due to image inpainting during the progressive generation process. The predefined cameras ${\{\bm{C}_i\}}_{i=0}^{N}$ are denoted by the extrinsic parameters $\bm{E}_i \in \mathbb{R}^{3 \times 4}$ and the shared intrinsic parameter $\bm{K}\in \mathbb{R}^{3 \times 3}$, where $N$ denotes the number of cameras. Based on the initial camera $\bm{C}_0$, 2D pixels are transformed to 3D space to construct the initial point cloud $\bm{P}_0$ through a series of geometric transformations $\mathcal{T}_{unproj}$: 
\begin{eqnarray}
 \bm{P}_0 = \mathcal{T}_{unproj}(\bm{I}_0, \bm{D}_0, \bm{E}_0, \bm{K}) .
\end{eqnarray}

After obtaining the initial point cloud $\bm{P}_0$, additional point clouds need to be merged into the existing ones at each camera pose. Specifically, at the $i^{th}$ camera, the existing 3D point cloud $\bm{P}_{i-1}$ is projected into 2D space through a series of geometric transformations $\mathcal{T}_{proj}$. Due to changes in camera pose, this projection produces a partial image $\hat{\bm{I}}_i$ and a mask $\hat{\bm{M}}_i$ indicating the area for inpainting:
\begin{eqnarray}
 \hat{\bm{I}}_i, \hat{\bm{M}}_i = \mathcal{T}_{proj}(\bm{P}_{i-1}, \bm{E}_i, \bm{K}) .
\end{eqnarray}

Then $F_{inpaint}$ is used to generate the image $\bm{I_i}$ based on $\hat{\bm{I}}_i$, $\hat{\bm{M}}_i$, and $y$, followed by $F_d$ to obtain the depth map $\bm{D}_i$ from $\bm{I_i}$. 
Since there is some difference between the depth maps of two neighboring frames, $\bm{D}_i$ needs to be processed by minimizing the difference between the overlapping regions of the two point clouds to get the aligned depth $\bm{D}_{i}^{a}$:
\begin{eqnarray}
 \bm{D}_{i}^{a} = {\bm{f}}_{a}(\mathcal{T}_{unproj}({\bm{I}_i, \bm{D}}_i, \bm{E}_i, \bm{K}), \bm{P}_{i-1}, \hat{\bm{M}}_i=1) , \label{eq:aligned_depth}
\end{eqnarray}
where the function ${\bm{f}}_{a}(\cdot)$ minimizes the difference between the overlapping parts $(\hat{\bm{M}}_i=1)$ of two point clouds.
Then the inpainted pixels $(\hat{\bm{M}}_i=0)$ of $\bm{I}_i$ need to be transformed to 3D space. The updated point cloud $\bm{P}_i$ is defined as:
\begin{equation}
  {\bm{P}}_i = {\bm{f}}_{u}({\bm{P}}_{i-1}, \mathcal{T}_{unproj}(\bm{I}_i, \bm{D}_{i}^{a}, \bm{E}_i, \bm{K}), \hat{\bm{M}}_i=0),
\end{equation}
where the function ${\bm{f}}_{u}(\cdot)$ merges the new point cloud into the existing point cloud $\bm{P}_{i-1}$. The above steps are repeated $N$ times to obtain the final point cloud $\bm{P}_N$.

\noindent\textbf{3D Gaussians Initialization.}
We use $\bm{P}_N$ as the initial SfM \cite{schonberger2016sfm} points to initialize 3DGS. Since the initial $(N+1)$ views are not sufficient to train the 3DGS to produce reasonable outputs, we choose to add additional $M$ support views to form the image training set $\bm{I}_{i=0}^{N+M}$ of 3DGS. Unlike previous work \cite{chung2023luciddreamer}, we choose to take the depth of the center of each depth map $\bm{D}_i$ as the radius of the spheres. The cameras are shifted $\pm 5^{\circ}$ along each sphere to get new cameras ${\{\bm{C}_i\}}_{i=N+1}^{N+M}$. The image training set $\bm{I}_{i=0}^{N+M}$ are obtained by reprojection from $\bm{P}_N$ using ${\{\bm{C}_i\}}_{i=0}^{N+M}$:
\begin{equation}
\label{depth}
 \bm{I}_i, \hat{\bm{M}}_i = \mathcal{T}_{proj}(\bm{P}_N, \bm{E}_i, \bm{K}),
\end{equation}
where $i \in \{0, ... , N+M\}$. When optimizing 3DGS, we only consider the valid image regions $(\hat{\bm{M}}_i=1)$ for the support views $\bm{I}_{i=N+1}^{N+M}$ to prevent 3DGS from learning the erroneous details of reprojection.

\subsection{Hierarchical Depth Prior-based Regularization}
3DGS represents the scene more realistically through numerous 3D Gaussians with geometric and appearance attributes. The scenes generated by 3DGS in the progressive scene generation framework tend to be ambiguous and artifactual since the scene contains millions of attributes of Gaussian distributions optimized only via gradient descent based on photometric loss.
Previous work \cite{yuan2024dreamscape, li2024dreamscene} utilizes score distillation to achieve 3D scenes with consistency, which improves the quality of novel view synthesis to some extent.
 Despite their progress, some limitations remain: (1) Lack of precise constraints on 3D cues and depth information in the optimization process. (2) Neglecting effective supervision of the visual and geometric smoothness of the scene. The above issues limit the realism and continuity of 3D scene generation.
To this end, we propose a hierarchical Depth Prior-based Regularization (DPR) mechanism that implements multi-level regularization on the 3D Gaussians utilizing high-quality depth prior. 
Specifically, we implement joint constraints on the depth maps generated by 3DGS at the pixel level and distribution level by utilizing the Huber loss and Central Moment Discrepancy (CMD), respectively. Furthermore, the bilateral filter is leveraged to enhance the continuity of the depth information. 
In the following, the depth map $\bm{D}$ is obtained by the monocular depth estimator $F_d$.
3DGS estimates the z-depth map $\hat{\bm{D}}$ of all pixel by the \Cref{z-depth}.

\noindent \textbf{Depth Estimation Accuracy Constraints.}
We utilize a multi-scale constraint paradigm at the pixel level and distribution level to achieve accurate estimation of depth information.
The depth of object edges is difficult to estimate and inaccurate in depth maps. The edges of objects tend to be regions with large image gradients. Thus, to apply more attention to the edges, we design a gradient-aware Huber-based depth loss for implementing pixel-level depth constraints and adaptive depth regularization, denoted as follows:
\begin{equation}
   \mathcal{L}_{DPR}^{pixel}=\left\{\begin{array}{l}
g_{\mathrm{rgb}} \frac{1}{|\hat{\bm{D}}|} \sum\|\bm{D}-\hat{\bm{D}}\|_1, \text { if }\|\bm{D}-\hat{\bm{D}}\|_1  > \delta \\
g_{\mathrm{rgb} \mid} \frac{1}{|\bm{D}|} \sum \frac{(\bm{D}-\hat{\bm{D}})^2+\delta^2}{2 \delta}, \text { otherwise }
\end{array}\right.,
\end{equation}
where $g_{rgb} = exp(- \nabla)$ and $\nabla$ is the gradient of the current aligned RGB image, $\delta=0.2\max\|\bm{D}-\hat{\bm{D}}\|_1$, and $|\hat{\bm{D}}|$ indicates the total number of pixels in $\hat{\bm{D}}$. 
Image edges with larger gradients are dynamically assigned smaller learning weights.
Constraining two depth maps only at the pixel level ignores the discrepancy of their distributions.
Therefore, we implement distribution-level alignment between depth maps based on CMD, which has been widely used in domain adaptation to estimate the discrepancy between two domains \cite{zellinger2019robust}. CMD can utilize higher-order moments to effectively capture higher-order statistical information without kernel function dependence. 
For a random variable $\bm{X}$, the $k$-th central moment is given by: $\mu_k=\mathbb{E}\left[(\bm{X}-\mathbb{E}[\bm{X}])^k\right]$, where $\mathbb{E}[\bm{X}]$ denotes the mean of $\bm{X}$. For two distributions $\bm{P}$ and $\bm{Q}$, the CMD computes the discrepancy by summing the differences of their corresponding central moments up to order $K$ :

\begin{equation}
   \mathcal{D}_{CMD}^K(\bm{P}, \bm{Q})=\sum_{k=1}^K\left\|\mu_k^P-\mu_k^Q\right\|_2,
\end{equation}
where $\mu_k^P$ and $\mu_k^Q$ are the $k$-th central moments of distributions $\bm{P}$ and $\bm{Q}$, respectively, and $\|\cdot\|_2$ represents the Euclidean norm.
The CMD-based depth loss is expressed as:
\begin{equation}
    \mathcal{L}_{DPR}^{dist} = \mathcal{D}_{CMD}^K(\bm{D}, \hat{\bm{D}}).
\end{equation}

\noindent \textbf{Depth Smoothness Constraints.}
To address the problem that object boundaries in 3DGS-rendered images often appear to have nonsmooth edges, we propose a depth loss based on the bilateral filter \cite{tomasi1998bilateral}.
Bilateral filtering is a typical nonlinear filtering method that simultaneously considers both the space and value domain information, allowing the removal of depth noise while preserving the boundaries and details of the image.
Given two pixels $p$ and $q$ in the depth map with coordinates $(i,j)$ and $(m,n)$ respectively. The spatial kernel and color kernel of bilateral filtering are denoted as:
\begin{equation}
\small
  \mathcal{L}_{DPR}^{smooth}=\frac{1}{|\mathcal{N}(p)|} \sum_{q \in \mathcal{N}(p)} \mathcal{G}_s(p, q) \cdot \mathcal{G}_c(p, q) \cdot(\hat{\bm{D}}_{p}-\hat{\bm{D}}_{q})^2,
\end{equation}
where $|\mathcal{N}(p)|$ is the number of pixels in the neighborhood of pixel $p$, $\hat{\bm{D}}_p$ is the depth value at pixel $p$, spatial kernel is denoted as $\mathcal{G}_s(p, q)=exp (-\frac{(i-m)^2+(j-n)^2}{2 \sigma_s^2})$, and color kernel is denoted as $\mathcal{G}_c(p, q)=exp(-\frac{\|\hat{\bm{D}}_p-\hat{\bm{D}}_q\|^2}{2 \sigma_c^2})$.
Consequently, the loss of DRP is expressed as:
\begin{equation}
    \mathcal{L}_{DPR} = \lambda_1\, \mathcal{L}_{DPR}^{pixel} + \lambda_2\,\mathcal{L}_{DPR}^{dist} + \lambda_3\,\mathcal{L}_{DPR}^{smooth},
\end{equation}
where  $\lambda_1$, $\lambda_2$, and $\lambda_3$ are set to $0.7$, $0.1$ and $1.0$, respectively.

\subsection{Structured Context-guided Compression}
The microscopic 3D Gaussians with optimizable geometric and appearance attributes in 3DGS make it a powerful advantage for rendering a variety of scenes. Nevertheless, a complex and larger-scale scene often requires a prohibitively large number of 3D Gaussians for fine-grained representation, resulting in significant storage overhead.
Furthermore, in real-world applications, low-cost and lightweight models are more conducive to deployment and rapid scene generation.
Due to the unorganized and sparse properties of 3D Gaussians \cite{chen2024survey}, compressing 3D Gaussians is a challenging task. Mainstream 3DGS compression methods mostly focus only on the ``values'' \cite{fan2023lightgaussian, navaneet2023compact3d}, ignoring the structural correlation between their 3D Gaussians, resulting in a large amount of structural redundancy and inefficient compression.
Scaffold-GS \cite{lu2024scaffold} introduces anchors to cluster nearby relevant 3D Gaussians and utilizes the anchors' properties to predict the 3D Gaussians' properties. Although Scaffold-GS exploits the spatial correlations among 3D Gaussians, the independence of anchors leads to a large number of sparse and disordered anchors that are difficult to compress.
HAC \cite{chen2025hac} models the relationship among the anchors to some extent, but it insufficiently quantifies the anchors, leading to sub-optimal storage compression results.
To take full advantage of the correlation between unorganized anchors, inspired by Scaffold-GS and HAC, we propose a Structured Context-guided Compression (SCC) mechanism that utilizes a structured hash feature mesh to model the context of the anchor attributes.

\noindent\textbf{Description of anchors.}
In Scaffold-GS, each anchor is composed of a location $\bm{x}^a \in \mathbb{R}^3$ and an anchor attribute $\mathcal{A}=\{\bm{f}^a \in \mathbb{R}^{D^a},\bm{l} \in \mathbb{R}^6, \bm{o} \in \mathbb{R}^{3K}\}$, where each component represents anchor feature, scaling, and offsets, respectively. 
During the rendering phase, the anchor feature is fed into the MLPs to generate attributes for 3D Gaussians, whose locations are determined by adding $\bm{x}^a$ and $\bm{o}$, where $\bm{l}$ is utilized to regularize both locations and shapes of the Gaussians.
The attributes inferred from the anchor attributes by neighboring 3D Gaussians should be similar. 
Thus, following the methodology of HAC, we utilize a structured hash grid to model the inherent spatial consistency of independent anchors.
The core idea is to use the hash feature $\bm{f}^h$, obtained by implementing trilinear interpolation in the hash grid, to model the context of anchor attributes.
There is rich mutual information between anchor feature $\bm{f}^a$ and hash feature $\bm{f}^h$, thus maximizing the conditional probability of both can reduce the entropy of the feature and bit consumption \cite{chen2025hac}.
\begin{figure*}[t]
\centering
\includegraphics[width=0.78\textwidth]{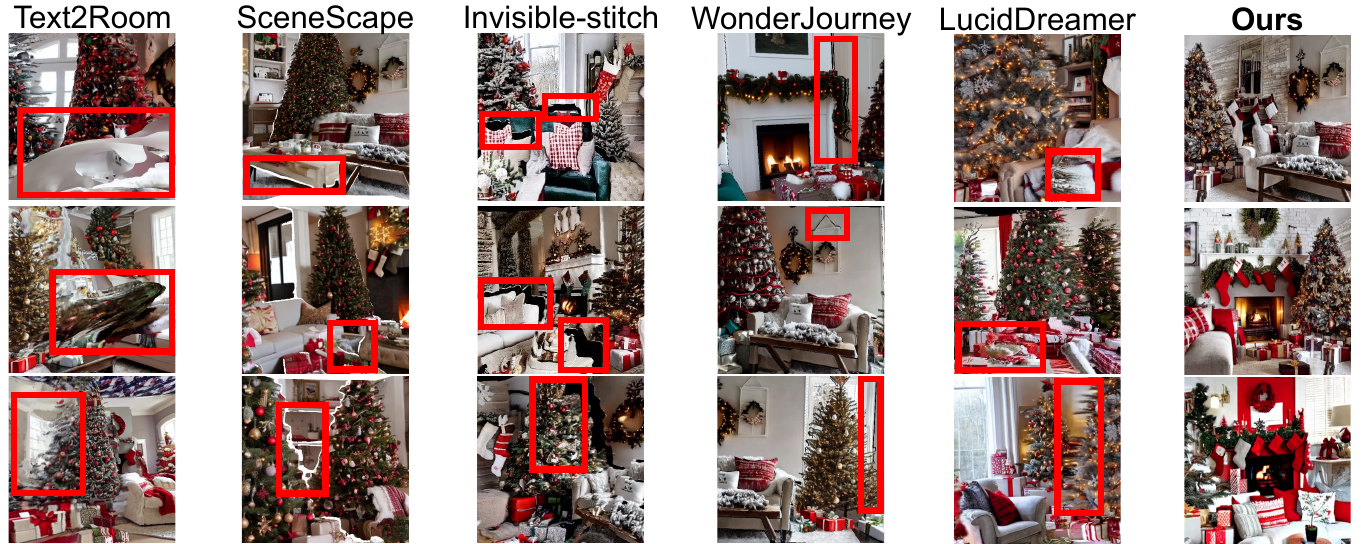}
\caption{Qualitative comparison results. The input text prompt is: ``A cozy living room in Christmas''}
\label{fig1}
\end{figure*}

\begin{figure*}[t]
\centering
\includegraphics[width=0.78\textwidth]{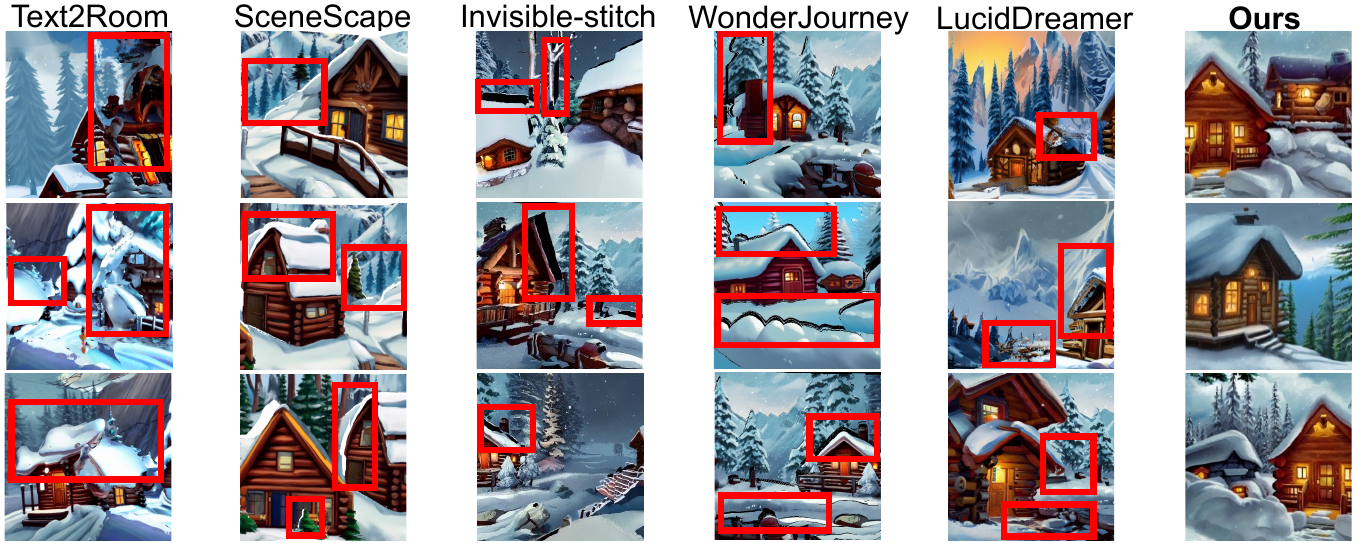}
\caption{Qualitative comparison results. The input text prompt is: ``A small cabin on top of a snowy mountain, Disney style''}
\label{fig2}
\end{figure*}
\begin{table*}
\centering
\renewcommand{\arraystretch}{1.0}
\setlength{\tabcolsep}{4pt}
\resizebox{0.8\textwidth}{!}{%
\begin{tabular}{cccccccc}
\toprule
\multirow{2}{*}{Models} & \multirow{2}{*}{Size (MB) $\downarrow$} & \multirow{2}{*}{CLIP-Score $\uparrow$} & \multicolumn{3}{c}{CLIP-IQA $\uparrow$}       & \multirow{2}{*}{BRISQUE $\downarrow$} & \multirow{2}{*}{NIQE $\downarrow$} \\ \cmidrule{4-6}
                        &                                       &                                        & Quality       & Colorful      & Sharp         &                                       &                                    \\ \midrule
Text2Room \cite{hollein2023text2room}              & 204.41                                & 29.45                                  & 0.60          & 0.77          & 0.34          & 27.24                                 & 3.43                               \\
SceneScape \cite{fridman2024scenescape}             & 189.00                                & 30.97                                  & 0.56          & 0.76          & 0.32          & 31.98                                 & 3.95                               \\
Invisible-stitch \cite{engstler2024invisible}       & 430.55                                & 31.16                                  & 0.63          & 0.68          & 0.41          & 26.19                                 & 3.56                               \\
WonderJourney \cite{yu2024wonderjourney}          & $-$                                & 30.76                                  & 0.58          & 0.77          & 0.38          & 27.46                                 & 3.47                               \\
LucidDreamer \cite{chung2023luciddreamer}           & 571.63                                & 31.19                                  & 0.66          & 0.77          & 0.42          & 24.07                                 & 3.05                               \\
\textbf{BloomScene (Ours)}           & \textbf{99.22}                        & \textbf{31.78}                         & \textbf{0.70} & \textbf{0.79} & \textbf{0.45} & \textbf{20.16}                        & \textbf{2.92}                      \\ \bottomrule
\end{tabular}%
}
\caption{Performance comparison among BloomScene and baselines. Our approach achieves the best results.}
\label{table1}
\end{table*}

\begin{table*}[]
\centering
\renewcommand{\arraystretch}{1.0}
\setlength{\tabcolsep}{6pt}
\resizebox{0.8\textwidth}{!}{%
\begin{tabular}{cccccccc}
\toprule
\multirow{2}{*}{Models} & \multirow{2}{*}{Size (MB) $\downarrow$} & \multirow{2}{*}{CLIP-Score $\uparrow$} & \multicolumn{3}{c}{CLIP-IQA $\uparrow$}       & \multirow{2}{*}{BRISQUE $\downarrow$} & \multirow{2}{*}{NIQE $\downarrow$} \\ \cmidrule{4-6}
                        &                                       &                                        & Quality       & Colorful      & Sharp         &                                       &                                    \\ \midrule

\textbf{BloomScene (full)}       & \textbf{99.22}  & \textbf{31.78}   & \textbf{0.70} & \textbf{0.79} &  \textbf{0.45} & \textbf{20.16}  & \textbf{2.92} \\
w/o DPR                          & 101.35                                 & 31.38                                   & 0.67                                 & 0.77                                 & 0.40                                 & 22.15                                  & 3.03                                 \\
w/o SCC                          & 569.33                                 & 31.60                                   & 0.68                                 & 0.78                                 & 0.44                                 & 22.46                                  & 2.96                                 \\
w/o $\mathcal{L}_{DPR}^{smooth}$ & 101.11                                 & 31.68                                   & 0.67                                 & 0.78                                 & 0.42                                 & 20.37                                  & 2.93                                 \\
w/o $\mathcal{L}_{DPR}^{dist}$   & 100.79                                 & 31.72                                   & 0.67                                 & 0.77                                 & 0.42                                 & 20.76                                  & 2.94                                 \\
w/o $\mathcal{L}_{DPR}^{pixel}$  & 101.19                                 & 31.54                                   & 0.66                                 & 0.77                                 & 0.42                                 & 20.57                                  & 2.95                                \\
 \bottomrule
\end{tabular}%
}
\caption{Ablation results of different components.}
\label{table2}
\end{table*}

\begin{figure}[t]
\centering
\includegraphics[width=1.0\columnwidth]{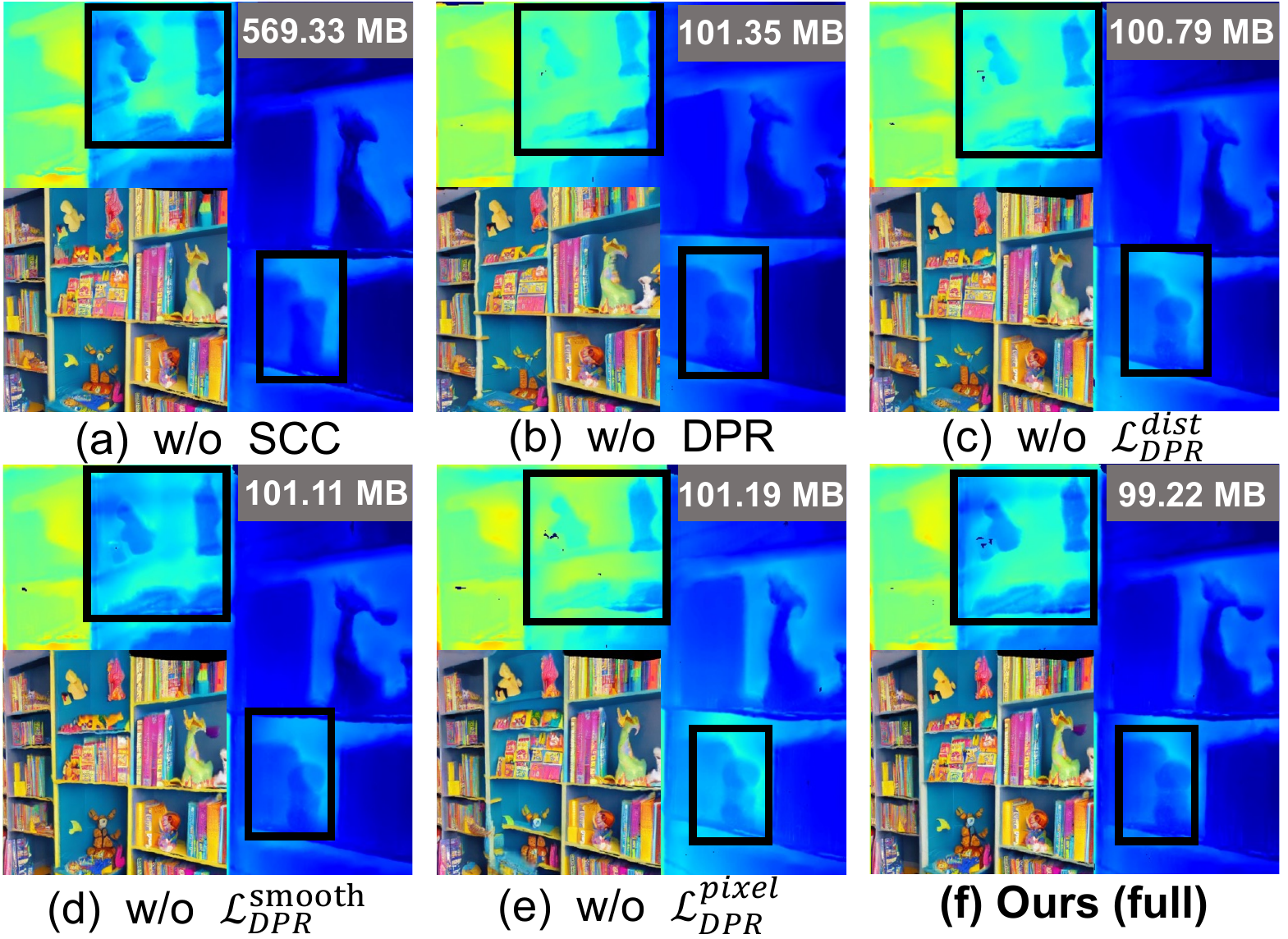}
\caption{Visualization of ablation results.}
\label{fig3}
\end{figure}

\noindent\textbf{Anchor feature modeling.}
To facilitate entropy coding, the values of $\bm{\mathcal{A}}$ must be quantized into a finite set. 
HAC utilizes uniform distribution-based noise and rounding operations to implement quantization during the training and testing phases, respectively, which are not sufficiently dynamic and smooth.
Therefore, we propose a dynamic quantization strategy. Specifically, for the $i$-th anchor $\bm{x}_i^a$, we denote $\bm{f}_i$ as any of its $\bm{\mathcal{A}}_i$'s components: $\bm{f}_i \in \{\bm{f}_i^a, \bm{l}_i, \bm{o}_i\} \in \mathbb{R}^D$, where $D \in \{D_a, 6, 3K\}$ is its respective dimension. 
In the training phase, we construct a Gaussian noise to update the features, denoted as:
\begin{equation}
    \hat{\bm{f}}_i  =\bm{f}_i+\mathcal{N}(0, \omega_i^2),
\end{equation}
where $\omega_i = \eta_i\,(1+\text{Tanh}(\mathcal{F}_q(\bm{f_i^h}))$ with $\eta_i \in \{2.5e-1, 2.5e-4, 5e-2\}$, and $\mathcal{F}_q$ is an MLP for generating factors to dynamically optimize quantization.
In the inference phase, we utilize a semi-soft rounding operation to make the quantized results closer to the true values, but still retain some discretization, expressed as:
\begin{equation}
    \hat{\bm{f}}_i=k \cdot \omega_i+\operatorname{Tanh}\left(\frac{\bm{f}_i-k \cdot \omega_i}{\tau}\right) \cdot \omega_i,
\end{equation}
where the smoothing hyperparameter $\tau$ is 1.
To measure and reduce the bit consumption of $\bm{f}_i$ during training, we need to estimate its probability in a microscopic manner. All three attributes of the anchors exhibit statistical tendencies of Gaussian distributions \cite{chen2025hac}. Thus, based on the independence of the anchor attributes, we construct Gaussian distributions for all anchor attributes, with $\mu$ and $\sigma$ in the respective distributions estimated by an MLP $\mathcal{F}_g$ from $\bm{f}^h$.
The probability of $\hat{\bm{f}}_i$ is computed as:

\begin{equation}
p\left(\hat{\bm{f}}_i\right)  =\int_{\hat{\bm{f}}_i-\frac{1}{2} \omega_i}^{\hat{\bm{f}}_i+\frac{1}{2} \omega_i} \phi_{\mu_i, \sigma_i}(x) d x,
\end{equation}
where $\phi$ represents the probability density function and $\mu_i, \sigma_i  =\mathcal{F}_g(\bm{f}_i^h)$.
Ultimately, we define the entropy loss as the sum of the bit consumption of all $\hat{\bm{f}}_i$:
\begin{equation}
    \mathcal{L}_{entropy}= \beta \sum_{\bm{f} \in\left\{\bm{f}^a, \bm{l}, \bm{o}\right\}} \sum_{i=1}^N \sum_{j=1}^D\left(-\log _2 p\left(\hat{\bm{f}}_{i}^j\right)\right),
\end{equation}
where $\beta = \frac{1}{N(D^a + 6 + 3K)}$, $N$ is the number of anchors, $D^a=50$ is the anchor feature dimension, $K=10$ is the number of learnable offsets and $\hat{\bm{f}}_{i}^j$ means the $j$-th  dimension value of $\hat{\bm{f}}_{i}$. 
Minimizing the entropy loss achieves a high probability estimation of $p(\hat{\bm{f}}_i)$ that guides the learning of the contextual model.
The SCC loss is denoted as:
\begin{equation}
    \mathcal{L}_{SCC} =  \lambda_4\,\mathcal{L}_{vol} + \lambda_5\,\mathcal{L}_{entropy},
\end{equation}
where $\lambda_4$ and $\lambda_5$ are set to to $1e-2$ and $2e-3$, $\mathcal{L}_{vol}$ is the regularization term defined in \cite{lu2024scaffold}.

\subsection{Optimization Objectives}
The final loss we use for optimization is defined as follows:
\begin{equation}
    \mathcal{L} =  \mathcal{L}_{RGB} + \mathcal{L}_{DPR} +   \mathcal{L}_{SCC},
\end{equation}
where $\mathcal{L}_{RGB}$ is the original photometric loss proposed in \cite{kerbl20233d}.

\section{Experiments}

\subsection{Text prompts and Evaluation Metrics}
\noindent\textbf{Text Prompts}. To achieve a fair and comprehensive comparison, we select 9 text prompts describing the indoor, outdoor, and artistic style scenes: (1) A living room with a lit furnace, couch and cozy curtains, bright lamps that make the room look well-lit. (2) A cozy living room in Christmas. (3) A small cabin on top of a snowy mountain, Disney style. (4) A suburban street in North Carolina on a bright, sunny day. (5) Simple museum, pictures, paintings, artistic, best quality, dimly lit. (6) A children's room filled with toys and books. (7) A sunroom with floor-to-ceiling windows overlooking the garden, comfortable chairs, and a coffee table inside. (8) A sunny beach with fine sand and blue water, with a backdrop of blue sky and white clouds. (9) A winter snow scene with snow-covered trees and houses.

\noindent\textbf{Evaluation Metric}. Previous reference-based metrics (\emph{e.g.}, PSNR and LPIPS \cite{zhang2018lpips}) are not suitable for this generation task due to the lack of 3D scenes related to text prompts as reference. Therefore, six 2D metrics are used to assess the quality of the generated scenes comprehensively. We use BRISQUE \cite{mittal2012brisque} and NIQE \cite{mittal2012niqe} for reference-free image quality assessment, and CLIP score \cite{hessel2021clipscore} to measure the alignment between the rendered images and the input text prompts. In addition, the appearance and feel of the images are evaluated in a way that is more closely aligned with human perception through the Colorful, Quality, and Sharp metrics of CLIP-IQA \cite{wang2023clipiqa}. Moreover, we use the size of the model to measure the storage overhead.

\subsection{Implementation Details}
To maximize the generalization ability of the proposed BloomScene, we use pre-trained models to build the entire architecture. Specifically, Stable Diffusion v1.5 \cite{rombach2022high} is used to generate the initial image from the text prompt. If the input is an image without a corresponding text description, LLaVa \cite{llava} is used to generate the corresponding text prompt from the image, constituting an image-text pair. We use the Stable Diffusion v1.5 Inpainting model \cite{rombach2022high} as the text-conditioned image inpainting model. We use ZoeDepth \cite{bhat2023zoedepth} as the monocular depth estimator. To generate 3D scenes, we move the camera with a rotation of 0.63 radians. All experiments are done on a single NVIDIA A800 GPU. All experimental results are averaged over multiple experiments using five different random seeds.

\subsection{Comparison with State-of-the-Art Methods}
\label{sec:comparison}
We compare the proposed BloomScene with five representative and reproducible methods, including progressive 3D scene generation methods: Text2Room \cite{hollein2023text2room}, Invisible-stitch \cite{engstler2024invisible} and LucidDreamer \cite{chung2023luciddreamer}, and perpetual view generation methods: SceneScape \cite{fridman2024scenescape} and WonderJourney \cite{yu2024wonderjourney}. We use the open-source codebase of the above models and modify the inputs to start from the same initial images and text prompts.

\noindent \textbf{Qualitative Results.}
We perform an intuitive qualitative analysis. We show the rendered RGB images of our method and baseline methods in the new viewpoints in Figure \ref{fig1} and Figure \ref{fig2}. We have the following observations: 
({\romannumeral1}) SceneScape, WonderJourney, and Invisible-stitch generate relatively complete scene content, but clear breaks and geometric distortions can be observed in boxed areas. 
({\romannumeral2}) Text2Room uses a polygonal mesh to represent the scene, but its mesh fusion threshold filtering scheme results in incomplete detection of stretched regions, leading to distorted and over-smoothing areas in the scene.
({\romannumeral3}) LucidDreamer is currently the most visually effective progressive scene generation method but suffers from artifacts and geometric distortions in boxed areas. 
({\romannumeral4}) In contrast, our method preserves the necessary scene structures, significantly reduces artifacts and geometric distortions, and provides high-quality and realistic rendered results.

\noindent \textbf{Quantitative Results.}
\Cref{table1} shows the average quantitative results for multiple scenes. We can conclude the following points: ({\romannumeral1}) Overall, our method generates much higher quality 3D scenes with significantly reduced storage overhead, significantly outperforming the baseline models.
({\romannumeral2}) The storage overhead of our generated scenes is 4.3x and 5.8x lower than Invisible-stitch \cite{engstler2024invisible} and LucidDreamer \cite{chung2023luciddreamer} using 3DGS. The storage overhead is also significantly reduced compared to Text2Room \cite{hollein2023text2room} and SceneScape \cite{fridman2024scenescape} using mesh. 
({\romannumeral3}) We achieve the best performance on all the 2D metrics. In addition, the BRISQUE and NIQE scores are 20.16 and 2.92, respectively, which are reduced by 16.2\% and 4.3\% compared to the optimal scores. This indicates that our method fully utilizes the geometric information of the scene, effectively reduces distortions of the scene, and enhances the alignment of the scenes with the input text prompts.

\subsection{Ablation Studies}
To verify the necessity of the different components, we perform comprehensive ablation experiments using the same set of text prompts. \Cref{fig3} shows the rendered results and \Cref{table2} shows the average quantitative results for multiple scenes. 
({\romannumeral1}) Firstly, DPR is removed from BloomScene. The decreased performance and the worse depth rendered results indicate that effective supervision of depth information and smoothness during optimization is crucial in the realism and continuity of 3D scenes. 
({\romannumeral2}) Moreover, we replace SCC with the original 3DGS. The dramatic increase in scene storage overhead indicates that compression for complex and larger-scale scenes is very necessary. 
({\romannumeral3}) Eventually, we remove the loss terms from DPR. The degraded and worse performance in depth map smoothness and accuracy indicate that the various loss items of DPR are necessary.

\section{Conclusion}
In this paper, we propose BloomScene, a lightweight structured 3D Gaussian splatting for crossmodal scene generation.
Specifically, a crossmodal progressive scene generation framework is proposed to incrementally generate coherent scenes.
Furthermore, we propose a hierarchical depth prior-based regularization mechanism that utilizes multi-level constraints on depth accuracy and smoothness to enhance the realism and continuity of the generated scenes.
Finally, we propose a structured context-guided compression mechanism that utilizes structured hash grids to model the context of unorganized anchor attributes, thus significantly reducing storage overhead.
Comprehensive qualitative and quantitative experiments across multiple scenarios show that the proposed framework has significant advantages over several baselines.
Our framework opens up more possibilities for future virtual reality applications.

\section{Acknowledgements}
This work was supported by National Key R\&D Program of China 2021ZD0113502.

\bibliography{main}

\begin{thebibliography}{41}
\providecommand{\natexlab}[1]{#1}

\bibitem[{Barron et~al.(2022)Barron, Mildenhall, Verbin, Srinivasan, and Hedman}]{barron2022mip}
Barron, J.~T.; Mildenhall, B.; Verbin, D.; Srinivasan, P.~P.; and Hedman, P. 2022.
\newblock Mip-nerf 360: Unbounded anti-aliased neural radiance fields.
\newblock In \emph{Proceedings of the IEEE/CVF conference on computer vision and pattern recognition}, 5470--5479.

\bibitem[{Berger et~al.(2014)Berger, Tagliasacchi, Seversky, Alliez, Levine, Sharf, and Silva}]{berger2014state}
Berger, M.; Tagliasacchi, A.; Seversky, L.~M.; Alliez, P.; Levine, J.~A.; Sharf, A.; and Silva, C.~T. 2014.
\newblock State of the art in surface reconstruction from point clouds.
\newblock In \emph{35th Annual Conference of the European Association for Computer Graphics, Eurographics 2014-State of the Art Reports}. The Eurographics Association.

\bibitem[{Bhat et~al.(2023)Bhat, Birkl, Wofk, Wonka, and M{\"u}ller}]{bhat2023zoedepth}
Bhat, S.~F.; Birkl, R.; Wofk, D.; Wonka, P.; and M{\"u}ller, M. 2023.
\newblock Zoedepth: Zero-shot transfer by combining relative and metric depth.
\newblock \emph{arXiv preprint arXiv:2302.12288}.

\bibitem[{Chen and Wang(2024)}]{chen2024survey}
Chen, G.; and Wang, W. 2024.
\newblock A survey on 3d gaussian splatting.
\newblock \emph{arXiv preprint arXiv:2401.03890}.

\bibitem[{Chen et~al.(2025)Chen, Wu, Lin, Harandi, and Cai}]{chen2025hac}
Chen, Y.; Wu, Q.; Lin, W.; Harandi, M.; and Cai, J. 2025.
\newblock Hac: Hash-grid assisted context for 3d gaussian splatting compression.
\newblock In \emph{European Conference on Computer Vision}, 422--438. Springer.

\bibitem[{Chung et~al.(2023)Chung, Lee, Nam, Lee, and Lee}]{chung2023luciddreamer}
Chung, J.; Lee, S.; Nam, H.; Lee, J.; and Lee, K.~M. 2023.
\newblock Luciddreamer: Domain-free generation of 3d gaussian splatting scenes.
\newblock \emph{arXiv preprint arXiv:2311.13384}.

\bibitem[{Contributors(2023)}]{llava}
Contributors, X. 2023.
\newblock XTuner: A Toolkit for Efficiently Fine-tuning LLM.
\newblock \url{https://github.com/InternLM/xtuner}.

\bibitem[{Engstler et~al.(2024)Engstler, Vedaldi, Laina, and Rupprecht}]{engstler2024invisible}
Engstler, P.; Vedaldi, A.; Laina, I.; and Rupprecht, C. 2024.
\newblock Invisible Stitch: Generating Smooth 3D Scenes with Depth Inpainting.
\newblock \emph{arXiv preprint arXiv:2404.19758}.

\bibitem[{Fan et~al.(2023)Fan, Wang, Wen, Zhu, Xu, and Wang}]{fan2023lightgaussian}
Fan, Z.; Wang, K.; Wen, K.; Zhu, Z.; Xu, D.; and Wang, Z. 2023.
\newblock Lightgaussian: Unbounded 3d gaussian compression with 15x reduction and 200+ fps.
\newblock \emph{arXiv preprint arXiv:2311.17245}.

\bibitem[{Fridman et~al.(2024)Fridman, Abecasis, Kasten, and Dekel}]{fridman2024scenescape}
Fridman, R.; Abecasis, A.; Kasten, Y.; and Dekel, T. 2024.
\newblock Scenescape: Text-driven consistent scene generation.
\newblock \emph{Advances in Neural Information Processing Systems}, 36.

\bibitem[{Hessel et~al.(2021)Hessel, Holtzman, Forbes, Bras, and Choi}]{hessel2021clipscore}
Hessel, J.; Holtzman, A.; Forbes, M.; Bras, R.~L.; and Choi, Y. 2021.
\newblock Clipscore: A reference-free evaluation metric for image captioning.
\newblock \emph{arXiv preprint arXiv:2104.08718}.

\bibitem[{H{\"o}llein et~al.(2023)H{\"o}llein, Cao, Owens, Johnson, and Nie{\ss}ner}]{hollein2023text2room}
H{\"o}llein, L.; Cao, A.; Owens, A.; Johnson, J.; and Nie{\ss}ner, M. 2023.
\newblock Text2room: Extracting textured 3d meshes from 2d text-to-image models.
\newblock In \emph{Proceedings of the IEEE/CVF International Conference on Computer Vision}, 7909--7920.

\bibitem[{Hou et~al.(2024)Hou, Li, Chen, Yang, Qian, and Zhang}]{hou2024sceneweaver}
Hou, X.; Li, M.; Chen, J.; Yang, D.; Qian, Z.; and Zhang, L. 2024.
\newblock SceneWeaver: Text-Driven Scene Generation with Geometry-aware Gaussian Splatting.
\newblock In \emph{The 16th Asian Conference on Machine Learning (Conference Track)}.

\bibitem[{Kerbl et~al.(2023)Kerbl, Kopanas, Leimk{\"u}hler, and Drettakis}]{kerbl20233d}
Kerbl, B.; Kopanas, G.; Leimk{\"u}hler, T.; and Drettakis, G. 2023.
\newblock 3D Gaussian Splatting for Real-Time Radiance Field Rendering.
\newblock \emph{ACM Trans. Graph.}, 42(4): 139--1.

\bibitem[{Lee and Chang(2022)}]{lee2022understanding}
Lee, H.-H.; and Chang, A.~X. 2022.
\newblock Understanding pure clip guidance for voxel grid nerf models.
\newblock \emph{arXiv preprint arXiv:2209.15172}.

\bibitem[{Li et~al.(2024)Li, Shi, Zhang, Wu, Liao, Wang, Lee, and Zhou}]{li2024dreamscene}
Li, H.; Shi, H.; Zhang, W.; Wu, W.; Liao, Y.; Wang, L.; Lee, L.-h.; and Zhou, P. 2024.
\newblock DreamScene: 3D Gaussian-based Text-to-3D Scene Generation via Formation Pattern Sampling.
\newblock \emph{arXiv preprint arXiv:2404.03575}.

\bibitem[{Lin et~al.(2023)Lin, Gao, Tang, Takikawa, Zeng, Huang, Kreis, Fidler, Liu, and Lin}]{lin2023magic3d}
Lin, C.-H.; Gao, J.; Tang, L.; Takikawa, T.; Zeng, X.; Huang, X.; Kreis, K.; Fidler, S.; Liu, M.-Y.; and Lin, T.-Y. 2023.
\newblock Magic3d: High-resolution text-to-3d content creation.
\newblock In \emph{Proceedings of the IEEE/CVF Conference on Computer Vision and Pattern Recognition}, 300--309.

\bibitem[{Lu et~al.(2024)Lu, Yu, Xu, Xiangli, Wang, Lin, and Dai}]{lu2024scaffold}
Lu, T.; Yu, M.; Xu, L.; Xiangli, Y.; Wang, L.; Lin, D.; and Dai, B. 2024.
\newblock Scaffold-gs: Structured 3d gaussians for view-adaptive rendering.
\newblock In \emph{Proceedings of the IEEE/CVF Conference on Computer Vision and Pattern Recognition}, 20654--20664.

\bibitem[{Mildenhall et~al.(2021)Mildenhall, Srinivasan, Tancik, Barron, Ramamoorthi, and Ng}]{mildenhall2021nerf}
Mildenhall, B.; Srinivasan, P.~P.; Tancik, M.; Barron, J.~T.; Ramamoorthi, R.; and Ng, R. 2021.
\newblock Nerf: Representing scenes as neural radiance fields for view synthesis.
\newblock \emph{Communications of the ACM}, 65(1): 99--106.

\bibitem[{Mittal, Moorthy, and Bovik(2012)}]{mittal2012brisque}
Mittal, A.; Moorthy, A.~K.; and Bovik, A.~C. 2012.
\newblock No-reference image quality assessment in the spatial domain.
\newblock \emph{IEEE Transactions on image processing}, 21(12): 4695--4708.

\bibitem[{Mittal, Soundararajan, and Bovik(2012)}]{mittal2012niqe}
Mittal, A.; Soundararajan, R.; and Bovik, A.~C. 2012.
\newblock Making a “completely blind” image quality analyzer.
\newblock \emph{IEEE Signal processing letters}, 20(3): 209--212.

\bibitem[{Mohammad~Khalid et~al.(2022)Mohammad~Khalid, Xie, Belilovsky, and Popa}]{mohammad2022clip}
Mohammad~Khalid, N.; Xie, T.; Belilovsky, E.; and Popa, T. 2022.
\newblock Clip-mesh: Generating textured meshes from text using pretrained image-text models.
\newblock In \emph{SIGGRAPH Asia 2022 conference papers}, 1--8.

\bibitem[{M{\"u}ller et~al.(2022)M{\"u}ller, Evans, Schied, and Keller}]{muller2022instant}
M{\"u}ller, T.; Evans, A.; Schied, C.; and Keller, A. 2022.
\newblock Instant neural graphics primitives with a multiresolution hash encoding.
\newblock \emph{ACM transactions on graphics (TOG)}, 41(4): 1--15.

\bibitem[{Munkberg et~al.(2022)Munkberg, Hasselgren, Shen, Gao, Chen, Evans, M{\"u}ller, and Fidler}]{munkberg2022extracting}
Munkberg, J.; Hasselgren, J.; Shen, T.; Gao, J.; Chen, W.; Evans, A.; M{\"u}ller, T.; and Fidler, S. 2022.
\newblock Extracting triangular 3d models, materials, and lighting from images.
\newblock In \emph{Proceedings of the IEEE/CVF Conference on Computer Vision and Pattern Recognition}, 8280--8290.

\bibitem[{Navaneet et~al.(2023)Navaneet, Meibodi, Koohpayegani, and Pirsiavash}]{navaneet2023compact3d}
Navaneet, K.; Meibodi, K.~P.; Koohpayegani, S.~A.; and Pirsiavash, H. 2023.
\newblock Compact3d: Compressing gaussian splat radiance field models with vector quantization.
\newblock \emph{arXiv preprint arXiv:2311.18159}.

\bibitem[{Ouyang et~al.(2023)Ouyang, Heal, Lombardi, and Sun}]{ouyang2023text2immersion}
Ouyang, H.; Heal, K.; Lombardi, S.; and Sun, T. 2023.
\newblock Text2immersion: Generative immersive scene with 3d gaussians.
\newblock \emph{arXiv preprint arXiv:2312.09242}.

\bibitem[{Poole et~al.(2022)Poole, Jain, Barron, and Mildenhall}]{poole2022dreamfusion}
Poole, B.; Jain, A.; Barron, J.~T.; and Mildenhall, B. 2022.
\newblock Dreamfusion: Text-to-3d using 2d diffusion.
\newblock \emph{arXiv preprint arXiv:2209.14988}.

\bibitem[{Radford et~al.(2021)Radford, Kim, Hallacy, Ramesh, Goh, Agarwal, Sastry, Askell, Mishkin, Clark et~al.}]{radford2021learning}
Radford, A.; Kim, J.~W.; Hallacy, C.; Ramesh, A.; Goh, G.; Agarwal, S.; Sastry, G.; Askell, A.; Mishkin, P.; Clark, J.; et~al. 2021.
\newblock Learning transferable visual models from natural language supervision.
\newblock In \emph{International conference on machine learning}, 8748--8763. PMLR.

\bibitem[{Ranftl et~al.(2020)Ranftl, Lasinger, Hafner, Schindler, and Koltun}]{ranftl2020towards}
Ranftl, R.; Lasinger, K.; Hafner, D.; Schindler, K.; and Koltun, V. 2020.
\newblock Towards robust monocular depth estimation: Mixing datasets for zero-shot cross-dataset transfer.
\newblock \emph{IEEE transactions on pattern analysis and machine intelligence}, 44(3): 1623--1637.

\bibitem[{Rombach et~al.(2022)Rombach, Blattmann, Lorenz, Esser, and Ommer}]{rombach2022high}
Rombach, R.; Blattmann, A.; Lorenz, D.; Esser, P.; and Ommer, B. 2022.
\newblock High-resolution image synthesis with latent diffusion models.
\newblock In \emph{Proceedings of the IEEE/CVF conference on computer vision and pattern recognition}, 10684--10695.

\bibitem[{Schonberger and Frahm(2016)}]{schonberger2016sfm}
Schonberger, J.~L.; and Frahm, J.-M. 2016.
\newblock Structure-from-motion revisited.
\newblock In \emph{Proceedings of the IEEE conference on computer vision and pattern recognition}, 4104--4113.

\bibitem[{Snavely, Seitz, and Szeliski(2006)}]{snavely2006photo}
Snavely, N.; Seitz, S.~M.; and Szeliski, R. 2006.
\newblock Photo tourism: exploring photo collections in 3D.
\newblock In \emph{ACM siggraph 2006 papers}, 835--846.

\bibitem[{Tang et~al.(2023)Tang, Ren, Zhou, Liu, and Zeng}]{tang2023dreamgaussian}
Tang, J.; Ren, J.; Zhou, H.; Liu, Z.; and Zeng, G. 2023.
\newblock Dreamgaussian: Generative gaussian splatting for efficient 3d content creation.
\newblock \emph{arXiv preprint arXiv:2309.16653}.

\bibitem[{Tomasi and Manduchi(1998)}]{tomasi1998bilateral}
Tomasi, C.; and Manduchi, R. 1998.
\newblock Bilateral filtering for gray and color images.
\newblock In \emph{Sixth international conference on computer vision (IEEE Cat. No. 98CH36271)}, 839--846. IEEE.

\bibitem[{Wang, Chan, and Loy(2023)}]{wang2023clipiqa}
Wang, J.; Chan, K.~C.; and Loy, C.~C. 2023.
\newblock Exploring clip for assessing the look and feel of images.
\newblock In \emph{Proceedings of the AAAI Conference on Artificial Intelligence}, volume~37, 2555--2563.

\bibitem[{Wang et~al.(2024)Wang, Lu, Wang, Bao, Li, Su, and Zhu}]{wang2024prolificdreamer}
Wang, Z.; Lu, C.; Wang, Y.; Bao, F.; Li, C.; Su, H.; and Zhu, J. 2024.
\newblock Prolificdreamer: High-fidelity and diverse text-to-3d generation with variational score distillation.
\newblock \emph{Advances in Neural Information Processing Systems}, 36.

\bibitem[{Yu et~al.(2024)Yu, Duan, Hur, Sargent, Rubinstein, Freeman, Cole, Sun, Snavely, Wu et~al.}]{yu2024wonderjourney}
Yu, H.-X.; Duan, H.; Hur, J.; Sargent, K.; Rubinstein, M.; Freeman, W.~T.; Cole, F.; Sun, D.; Snavely, N.; Wu, J.; et~al. 2024.
\newblock Wonderjourney: Going from anywhere to everywhere.
\newblock In \emph{Proceedings of the IEEE/CVF Conference on Computer Vision and Pattern Recognition}, 6658--6667.

\bibitem[{Yuan et~al.(2024)Yuan, Yang, Zhao, and Huang}]{yuan2024dreamscape}
Yuan, X.; Yang, H.; Zhao, Y.; and Huang, D. 2024.
\newblock DreamScape: 3D Scene Creation via Gaussian Splatting joint Correlation Modeling.
\newblock \emph{arXiv preprint arXiv:2404.09227}.

\bibitem[{Zellinger et~al.(2019)Zellinger, Moser, Grubinger, Lughofer, Natschl{\"a}ger, and Saminger-Platz}]{zellinger2019robust}
Zellinger, W.; Moser, B.~A.; Grubinger, T.; Lughofer, E.; Natschl{\"a}ger, T.; and Saminger-Platz, S. 2019.
\newblock Robust unsupervised domain adaptation for neural networks via moment alignment.
\newblock \emph{Information Sciences}, 483: 174--191.

\bibitem[{Zhang et~al.(2024)Zhang, Li, Wan, Wang, and Liao}]{zhang2024text2nerf}
Zhang, J.; Li, X.; Wan, Z.; Wang, C.; and Liao, J. 2024.
\newblock Text2nerf: Text-driven 3d scene generation with neural radiance fields.
\newblock \emph{IEEE Transactions on Visualization and Computer Graphics}.

\bibitem[{Zhang et~al.(2018)Zhang, Isola, Efros, Shechtman, and Wang}]{zhang2018lpips}
Zhang, R.; Isola, P.; Efros, A.~A.; Shechtman, E.; and Wang, O. 2018.
\newblock The unreasonable effectiveness of deep features as a perceptual metric.
\newblock In \emph{Proceedings of the IEEE conference on computer vision and pattern recognition}, 586--595.

\end{thebibliography}

\end{document}